\documentclass[conference]{IEEEtran}
\IEEEoverridecommandlockouts
\usepackage{cite}
\usepackage{amsmath,amssymb,amsfonts}
\usepackage{algorithmic}
\usepackage{graphicx}
\usepackage{textcomp}
\usepackage{xcolor}

\usepackage[hyphens]{url}
\usepackage{algorithm}
\usepackage{booktabs}
\usepackage{amssymb}
\usepackage{makecell}
\usepackage{multirow}

\def\BibTeX{{\rm B\kern-.05em{\sc i\kern-.025em b}\kern-.08em
    T\kern-.1667em\lower.7ex\hbox{E}\kern-.125emX}}
\begin{document}

\title{ELiTe: Efficient Image-to-LiDAR Knowledge Transfer for Semantic Segmentation
\thanks{*Corresponding author.}}

\author{\IEEEauthorblockN{Zhibo Zhang}
\IEEEauthorblockA{\textit{School of Computer Science} \\
\textit{Fudan University}\\
Shanghai, China \\
zhibozhang21@m.fudan.edu.cn}
\and
\IEEEauthorblockN{Ximing Yang}
\IEEEauthorblockA{\textit{School of Computer Science} \\
\textit{Fudan University}\\
Shanghai, China \\
xmyang19@fudan.edu.cn}
\and
\IEEEauthorblockN{Weizhong Zhang}
\IEEEauthorblockA{\textit{School of Data Science} \\
\textit{Fudan University}\\
Shanghai, China \\
zhangweizhongzju@gmail.com}
\and
\IEEEauthorblockN{Cheng Jin*}
\IEEEauthorblockA{\textit{School of Computer Science} \\
\textit{Fudan University}\\
Shanghai, China \\
jc@fudan.edu.cn}
}

\maketitle

\begin{abstract}
    Cross-modal knowledge transfer enhances point cloud representation learning in LiDAR semantic segmentation. Despite its potential, the \textit{weak teacher challenge} arises due to repetitive and non-diverse car camera images and sparse, inaccurate ground truth labels. To address this, we propose the Efficient Image-to-LiDAR Knowledge Transfer (ELiTe) paradigm. ELiTe introduces Patch-to-Point Multi-Stage Knowledge Distillation, transferring comprehensive knowledge from the Vision Foundation Model (VFM), extensively trained on diverse open-world images. This enables effective knowledge transfer to a lightweight student model across modalities. ELiTe employs Parameter-Efficient Fine-Tuning to strengthen the VFM teacher and expedite large-scale model training with minimal costs. Additionally, we introduce the Segment Anything Model based Pseudo-Label Generation approach to enhance low-quality image labels, facilitating robust semantic representations. Efficient knowledge transfer in ELiTe yields state-of-the-art results on the SemanticKITTI benchmark, outperforming real-time inference models. Our approach achieves this with significantly fewer parameters, confirming its effectiveness and efficiency.
\end{abstract}

\begin{IEEEkeywords}
3D point cloud, scene understanding, semantic segmentation, knowledge distillation
\end{IEEEkeywords}

\section{Introduction}

LiDAR semantic segmentation in autonomous driving relies on recent deep learning advancements \cite{DBLP:journals/tits/GaoPLGZ22}. Cross-modal knowledge transfer \cite{DBLP:conf/eccv/XuYGWYZZVKT22, DBLP:conf/eccv/XuGZZZCL22, DBLP:conf/aaai/XingYWYC23} enhances representation learning by leveraging semantic information from LiDAR and other modalities. However, the performance gap between teacher and student models in cross-modal transfer limits effectiveness due to potential weaknesses in the teacher model.

\begin{figure}[htbp]
\centering
\includegraphics[width=\columnwidth]{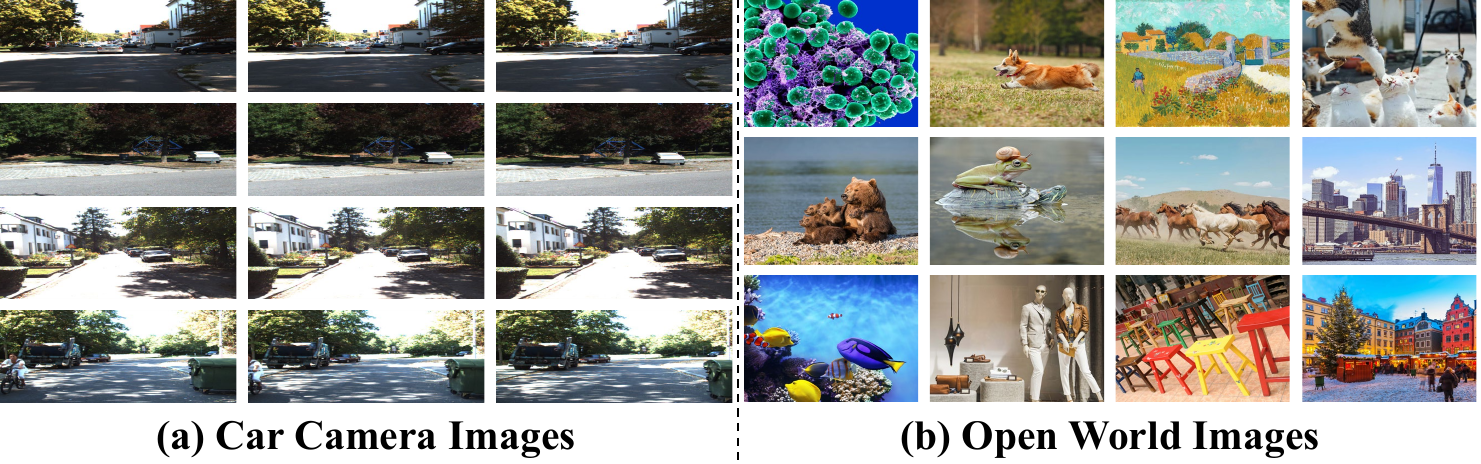}
\caption{Car Camera Images and Open World Images.}
\label{fig1}
\end{figure}

One reason is the data characteristic of the car-mounted camera images obtained during LiDAR data collection \cite{DBLP:conf/iccv/BehleyGMQBSG19}. These images usually depict the scene from a single viewpoint, resulting in high repeatability, limited diversity, and a monotonous style, as shown in Figure \ref{fig1}.
Additionally, there are often significant disparities in terms of dataset sizes. For instance, the SAM \cite{DBLP:conf/iccv/KirillovMRMRGXW23} is trained on a dataset consisting of approximately 11M images, while the widely used LiDAR dataset KITTI \cite{DBLP:conf/cvpr/GeigerLU12} contains only about 19K image frames.
Therefore, compared to models trained on many open images, training strong teacher models on car camera images becomes challenging because they are difficult to obtain comprehensive representations.

\begin{figure*}[htbp]
\centerline{\includegraphics[width=\textwidth]{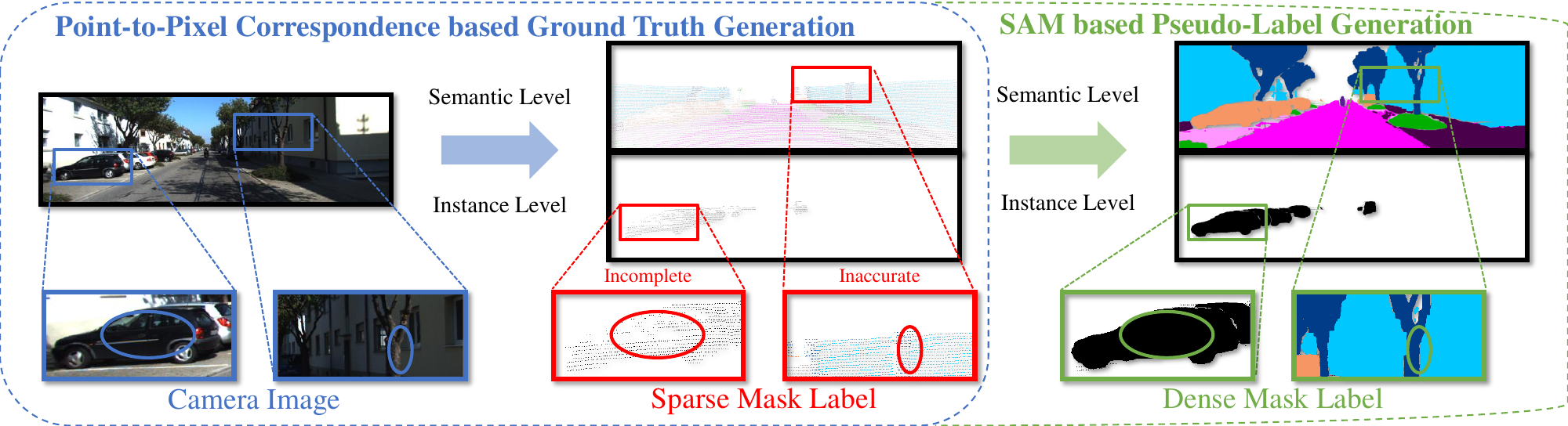}}
\caption{Existed one-stage Point-to-Pixel Correspondence based Ground Truth Generation (\textbf{PPC-GTG}) and our two-stage SAM-based Pseudo-Label Generation (\textbf{SAM-PLG}). The sparse mask labels generated by one-stage PPC-GTG are incomplete, inaccurate, and low-quality. The dense mask labels generated by two-stage SAM-PLG are more accurate, complete, and high quality.}
\label{fig2}
\end{figure*}

The weak supervision in training teacher models is another issue. Current methods \cite{DBLP:conf/eccv/XuGZZZCL22} use sparse mapping of images to LiDAR through perspective projection, creating 2D ground truth. This sparse mapping leads to weak supervision for teacher models, as seen in Figure \ref{fig2}. The lack of dense and precise semantic labels makes it challenging for the teacher model to learn semantic representations effectively.

In addition, the promising performance achieved by cross-modal knowledge transfer based methods \cite{DBLP:conf/eccv/XuGZZZCL22, DBLP:conf/cvpr/HouZMLL22} are always accompanied by a notable increase in the model size over the single-modal LiDAR semantic segmentation approaches \cite{DBLP:conf/cvpr/Zhu0WHM00L21, DBLP:conf/iccv/XuZDZSP21, DBLP:conf/cvpr/HouZMLL22}, in which modern neural networks with massive parameters have already been used. Therefore, the efficient optimization of large-scale parameters during the training process naturally emerges as a significant challenge.

To overcome the mentioned limitations, we introduce an \textbf{E}fficient image-to-\textbf{Li}DAR knowledge \textbf{T}ransf\textbf{e}r (ELiTe) paradigm for semantic segmentation. ELiTe incorporates Patch-to-Point Multi-Stage Knowledge Distillation (PPMSKD), a key component facilitating knowledge transfer from a Vision Foundation Model (VFM), specifically the Segment Anything Model (SAM) \cite{DBLP:conf/iccv/KirillovMRMRGXW23}. ELiTe employs Parameter-Efficient Fine-Tuning (PEFT) \cite{DBLP:conf/iclr/HuSWALWWC22} to expedite large-scale model training of VFM and enhance the teacher model.

To address weak supervision caused by sparse, incomplete, and inaccurate labels, ELiTe introduces an efficient semantic pseudo-label generation strategy, SAM-based Pseudo-Label Generation (SAM-PLG). SAM-PLG effectively converts sparse labels into dense and accurate ones, as depicted in Figure \ref{fig2}. Notably, SAM-PLG exhibits appealing features:
\begin{enumerate}
    \item At the instance level, it compensates for missing sparse mask labels on the car window area;
    \item At the semantic level, it produces clearer boundaries compared to sparse labels that exhibit inaccurate overlap between the tree and the building behind it;
    \item Dense labels provide richer supervisory information, mitigating the impact of inaccurate ground truth annotations derived from point-to-pixel correspondence.
\end{enumerate}
The resulting dense pseudo-label transforms weak supervision into strong supervision, reducing training difficulty and enhancing the teacher model's semantic representation learning.

ELiTe significantly improves LiDAR semantic segmentation using multi-modal data, particularly LiDAR point clouds. It achieves impressive performance even with a lightweight model for real-time inference, showcasing strong competitiveness on the SemanticKITTI benchmark.

Our main contributions can be summarized as follows:
\begin{itemize}
\item Introducing ELiTe: an Effective Image-to-LiDAR Knowledge Transfer paradigm using PPMSKD and PEFT. This enhances the teacher model's encoding ability, addressing performance disparity in existing studies between teacher and student models.
\item Our strategy, SAM-PLG, effectively transforms low-quality sparse labels into high-quality dense labels, mitigating the weak supervision issue and improving the teacher model's encoder capability.
\item ELiTe demonstrates state-of-the-art results on the SemanticKITTI benchmark, achieving real-time inference efficiency with significantly fewer parameters.
\end{itemize}

\section{Related Work}

\subsection{LiDAR Point Cloud Semantic Segmentation}
Recent research emphasizes \textbf{voxel-based methods} for efficiency and effectiveness. For example, Cylinder3D \cite{DBLP:conf/cvpr/Zhu0WHM00L21} uses cylindrical voxels and an asymmetrical network for better performance. \textbf{Multi-representation methods} are also trending, combining points, projection images, and voxels, and fusing features across branches. SPVNAS \cite{DBLP:conf/eccv/TangLZLLWH20} deploys point-voxel fusion with point-wise MLPs and NAS for architecture optimization. RPVNet \cite{DBLP:conf/iccv/XuZDZSP21} fuses range-point-voxel representations. Point-Voxel-KD \cite{DBLP:conf/cvpr/HouZMLL22} transfers knowledge from point and voxel levels to distill from a larger teacher to a compact student network. Notably, these methods consider only sparse LiDAR data, omitting appearance and texture cues from camera images. \textbf{Multi-modal fusion methods} have emerged to leverage the strengths of both cameras and LiDAR, combining information from these complementary sensors. RGBAL \cite{DBLP:conf/itsc/MadawiRSNKY19} maps RGB images using a polar-grid representation and fuses them at different levels. PMF \cite{DBLP:conf/iccv/ZhuangLJWLT21} collaboratively fuses data in camera coordinates. However, these methods require multi-sensor inputs during training and inference, posing computational challenges. Obtaining paired multi-modal data is also impractical in real-world scenarios.

Most methods aim for models with growing parameters, raising training costs. Ongoing efforts focus on compressing inference model parameters, but achieving practical real-time inference speed remains challenging.

\subsection{2D-to-3D Transfer Learning}
Transfer learning aims to leverage knowledge from data-rich domains to assist learning in data-scarce domains. The idea of distilling knowledge from 2D to 3D has been widely explored in the 3D community. While 2DPASS \cite{DBLP:conf/eccv/XuGZZZCL22} enhances semantic segmentation performance through 2D prior-related knowledge distillation, its teacher model is weak, and knowledge transfer is inefficient. In general vision tasks, pretraining significantly benefits various downstream tasks. Image2Point \cite{DBLP:conf/eccv/XuYGWYZZVKT22} transfers knowledge by replicating or inflating weights pre-trained on ImageNet \cite{DBLP:conf/cvpr/DengDSLL009} to a similar architecture of 3D models, efficiently adapting the domain by fine-tuning only specific layers. PointCLIP \cite{DBLP:conf/cvpr/ZhangGZLM0QG022} transfers representations learned from 2D images to different 3D point clouds but limits knowledge transfer to simple tasks like object classification due to reliance on point cloud projection for modal alignment.

Our work prioritizes real-time, large-scale LiDAR segmentation with minimal parameters and high component decoupling for efficient handling of complex challenges.

\section{Method}

\begin{figure*}[htbp]
\centerline{\includegraphics[width=\textwidth]{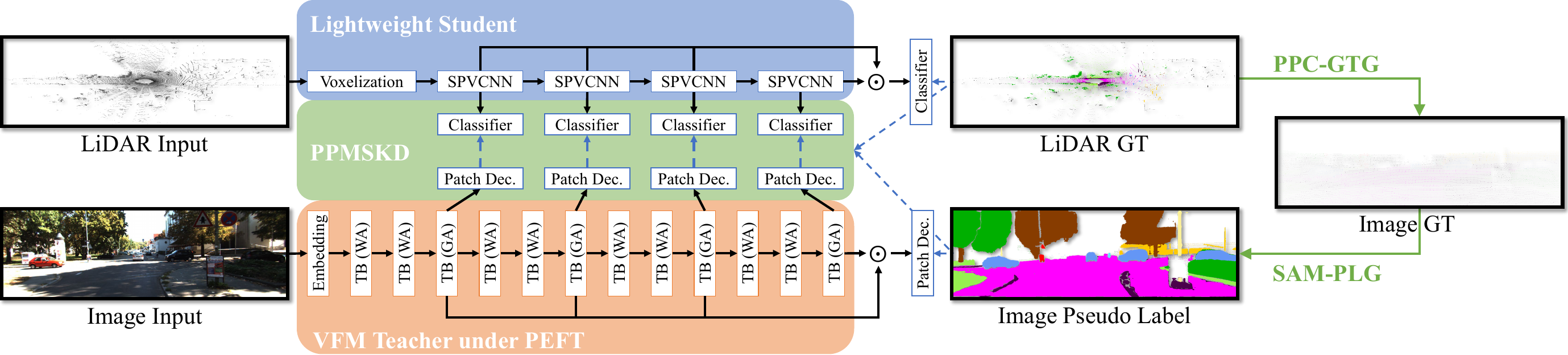}}
\caption{Framework Overview. It comprises three main components: VFM teacher, lightweight student, and Patch-to-Point Multi-Stage Knowledge Distillation(PPMSKD) networks. The teacher and student networks process image and LiDAR inputs, extracting multi-stage features. In the PPMSKD network, the knowledge from the teacher is transferred to the student. The VFM teacher network undergoes domain-adaptive fine-tuning via PEFT and is supervised by pseudo-labels generated by SAM-PLG. In this figure, TB(WA) and TB(GA) denote Transformer Blocks employing window and global attention, respectively, "Patch Dec." signifies the Patch Decoder, and $\odot$ represents concatenation. Solid lines delineate the data flow, while dashed lines represent the backpropagation supervisory signal.}
\label{fig3}
\end{figure*}

In this section, we outline our proposed method, ELiTe, including the teacher-student model, cross-modal knowledge distillation (PPMSKD), VFM fine-tuning (PEFT), and pseudo-label generation (SAM-PLG).

\subsection{Framework of ELiTe}
This paper proposes ELiTe, an efficient image-to-LiDAR knowledge transfer paradigm for semantic segmentation. ELiTe leverages rich color and texture information from image knowledge. The workflow includes three main components: the teacher network, the student network, and the distillation network. The student network uses PPMSKD in the distillation network to acquire image domain knowledge from the teacher network. The teacher network undergoes domain-adaptive fine-tuning with PEFT and is supervised by pseudo-labels from SAM-PLG. During inference, both the teacher and distillation networks are discarded to avoid additional computational burden in practical applications.

\subsection{VFM Teacher and Lightweight Student}
In Figure \ref{fig3}, both the VFM teacher network and the lightweight student network are utilized to encode both the image and the LiDAR point cloud. The teacher network utilizes the Vision Transformer (ViT) encoder \cite{DBLP:conf/iclr/DosovitskiyB0WZ21}, trained with SAM \cite{DBLP:conf/iccv/KirillovMRMRGXW23}, which leverages attention mechanisms. Attention \cite{DBLP:conf/nips/VaswaniSPUJGKP17}, proven effective across various vision tasks\cite{DBLP:conf/aaai/ZhangYWJ22, DBLP:conf/mm/CaoZCYJ22, DBLP:conf/aaai/ZhangYWJ22, DBLP:conf/icmcs/WangFWXZ23, wang2024residual, yang2021generate}, is a key component of ViT.

Within these networks, $L$ feature maps are derived from various stages to extract 2D features $\{F^{Patch}_{l}\}^L_{l=1}$ and 3D features $\{F^{Point}_{l}\}^L_{l=1}$. Single-stage features are utilized for knowledge distillation, whereas the concatenation of multi-stage features contributes to obtaining the final prediction scores.

\subsection{Patch-to-Point Multi-Stage Knowledge Distillation}
Our PPMSKD is a cross-modal knowledge transfer technique and its patch-to-point distillation scheme enables us to transfer rich knowledge from SAM trained on massive open-world images to LiDAR semantic segmentation. As shown in Figure \ref{fig3}, it comprises $L$ individual single-stage decoders used for both the teacher and student networks. These decoders restore downsampled feature maps to their original sizes and obtain prediction scores for knowledge distillation.

In LiDAR networks, \textbf{point decoders} are standard classifiers. $L$ point-level single-stage features are upsampled to the original point cloud size and input into single-stage classifiers to obtain prediction scores $\{S^{Point}_{l}\}^L_{l=1}$.

Within the teacher network based on the ViT architecture, the patch-level image features are extracted from the output of $L$ global attention blocks. In \textbf{patch decoders}, patch-level features are initially upsampled and feature dimensionality reduced through an upscaling layer, and then further upsampled to the original image size through interpolation. Finally, through pixel-to-point mapping, we can obtain pixels corresponding to LiDAR point clouds, thereby acquiring pixel-level predictions denoted as $\{S^{Patch}_{l}\}^L_{l=1}$, which correspond one-to-one with point-level features.

Key to knowledge transfer, PPMSKD enhances 3D representation at each stage through \textbf{multi-stage knowledge distillation} with auxiliary VFM priors. PPMSKD aligns image features with point cloud features, ensuring improved point cloud features. Vanilla KL divergence is the core distillation loss ($L_{KD}$), with multi-stage distillation losses as follows:
$$L_{KD}=\sum^L_{l=1}{D_{KL}(S^{Patch}_l||S^{Point}_l)}.$$

\begin{table*}[htbp]
\caption{LiDAR semantic segmentation results on the SemanticKITTI test benchmark}
\setlength\tabcolsep{4pt}
\begin{center}
\begin{tabular}{c|c|c|c|c|ccccccccccccccccccc}
\toprule
\textbf{Methods}
& \textbf{mIoU}
& \textbf{FPS(Hz)} 
& \rotatebox{90}{\textbf{\#Inference Params}}
& \rotatebox{90}{\textbf{\#Trainable Params}}
& \rotatebox{90}{car}
& \rotatebox{90}{bicycle}
& \rotatebox{90}{motorcycle}
& \rotatebox{90}{truck}
& \rotatebox{90}{other-vehicle}
& \rotatebox{90}{person}
& \rotatebox{90}{bicyclist}
& \rotatebox{90}{motorcyclist}
& \rotatebox{90}{road}
& \rotatebox{90}{parking}
& \rotatebox{90}{sidewalk}
& \rotatebox{90}{other-ground}
& \rotatebox{90}{building}
& \rotatebox{90}{fence}
& \rotatebox{90}{vegetation}
& \rotatebox{90}{trunk}
& \rotatebox{90}{terrain}
& \rotatebox{90}{pole}
& \rotatebox{90}{traffic-sign} \\
\midrule
TORNADONet \cite{DBLP:conf/icra/GerdzhevRTL21} & 63.1 & 4 & - & - & 94 & 56 & 48 & 40 & 38 & 64 & 60 & 35 & 90 & 66 & 75 & 29 & 91 & 66 & 86 & 67 & \textbf{72} & 58 & 66 \\\
SPVNAS \cite{DBLP:conf/eccv/TangLZLLWH20} & 66.4 & 3 & 12.5M & - & 97 & 52 & 51 & 60 & 59 & 66 & 65 & 44 & 90 & 68 & 75 & 17 & 91 & 66 & 86 & 73 & 71 & 64 & 67 \\
Cylinder3D \cite{DBLP:conf/cvpr/Zhu0WHM00L21} & 67.8 & 7 & 55.9M & 55.9M & 97 & 68 & 64 & 59 & 59 & 74 & 68 & 36 & 91 & 65 & 76 & 32 & 91 & 67 & 85 & 72 & 69 & 63 & 66 \\
AF2S3Net \cite{DBLP:conf/cvpr/ChengRTLL21} & 69.7 & - & - & - & 95 & 65 & \textbf{87} & 39 & 41 & \textbf{81} & 80 & \textbf{74} & 91 & 69 & 73 & \textbf{54} & 88 & 63 & 70 & 69 & 54 & 62 & \textbf{71} \\
RPVNet \cite{DBLP:conf/iccv/XuZDZSP21} & 70.3 & 5 & 24.8M & 24.8M & \textbf{98} & 68 & 69 & 44 & 61 & 76 & 74 & 73 & \textbf{93} & 70 & \textbf{81} & 33 & \textbf{94} & 72 & \textbf{87} & \textbf{75} & \textbf{72} & 65 & 61 \\
2DPASS-large \cite{DBLP:conf/eccv/XuGZZZCL22} & \textbf{72.9} & 7 & 45.6M & 74.3M & 97 & 64 & 74 & \textbf{74} & 62 & 78 & \textbf{81} & \textbf{74} & 90 & 67 & 75 & 40 & \textbf{94} & \textbf{73} & 86 & 61 & 71 & 65 & 70 \\
\midrule
RangeNet++\cite{DBLP:conf/iros/MiliotoVBS19} & 52.2 & 12 & 50.4M & 50.4M & 91 & 26 & 34 & 26 & 23 & 38 & 39 & 5 & 92 & 65 & 75 & 28 & 87 & 59 & 81 & 55 & 65 & 48 & 56 \\
PolarNet\cite{DBLP:conf/cvpr/0035ZDYXGF20} & 54.3 & 16 & 14.0M & 14.0M & 84 & 40 & 30 & 23 & 29 & 43 & 40 & 6 & 91 & 62 & 74 & 22 & 90 & 61 & 84 & 66 & 68 & 52 & 58 \\
Lite-HDSeg \cite{DBLP:conf/icra/RazaniCTL21} & 63.8 & 20 & - & - & 92 & 40 & 54 & 38 & 40 & 59 & 72 & 54 & 93 & 68 & 78 & 29 & 92 & 65 & 78 & 66 & 65 & 60 & 68 \\
2DPASS-base (Baseline) & 67.7 & \textbf{24} & \textbf{1.9M} & 26.5M & 97 & 62 & 60 & 55 & 60 & 75 & 80 & 22 & 90 & 62 & 75 & 30 & 91 & 66 & 86 & 72 & 70 & 63 & 69\\
Point-Voxel-KD \cite{DBLP:conf/cvpr/HouZMLL22} & 71.2 & 13 & 14.1M & - & 97 & 68 & 69 & 54 & 60 & 75 & 74 & 51 & 92 & \textbf{71} & 78 & 41 & 92 & 69 & \textbf{87} & 74 & \textbf{72} & 65 & 66 \\
\textbf{ELiTe (Ours)} & $\underline{71.4}$ & \textbf{24} & \textbf{1.9M} & \textbf{6.8M} & 97 & \textbf{69} & 67 & 58 & \textbf{66} & 78 & \textbf{81} & 44 & 90 & 66 & 75 & 35 & 92 & 69 & 86 & 74 & 70 & \textbf{66} & 70 \\
\bottomrule
\multicolumn{24}{l}{$^{\mathrm{*}}$Listed are non-real-time methods (\textbf{upper}) and real-time inference methods (\textbf{lower}).}
\end{tabular}
\label{tab:comparison}
\end{center}
\end{table*}

\subsection{Parameter-Efficient Fine-Tuning for VFM Teacher}

In fine-tuning large models, common choices include Full Fine-Tuning, Linear Probing, Robust Fine-Tuning \cite{DBLP:conf/cvpr/WortsmanIKLKRLH22, zhang2024robust}, and PEFT \cite{DBLP:conf/iclr/HuSWALWWC22, zhang2023adaptive}.

Recently, PEFT in NLP efficiently leverages pre-trained models for downstream applications, reducing the need for extensive fine-tuning and significantly cutting computational and storage demands. Notably, PEFT achieves comparable performance to full fine-tuning, optimizing cost-efficiency.

Given a pre-trained weight matrix $W(0)$, LoRA \cite{DBLP:conf/iclr/HuSWALWWC22}, a PEFT technique compatible with ViT, parameterizes $\Delta$ as a low-rank matrix by the product of two much smaller matrices:

$$W=W^{(0)}+\Delta=W^{(0)}+BA$$
where $W^{(0)}, \Delta\in R^{d_1\times d_2}, A\in R^{r\times d_2}$ and $B\in R^{d_1\times r}$ with $r\ll\{d_1,d_2\}$. During fine-tuning, only
$A$ and $B$ are updated. 

We chose to employ AdaLoRA \cite{zhang2023adaptive} for facilitating domain adaptation within the context of VFM teacher. AdaLoRA subsequently allocates LoRA parameter budgets adaptively based on importance scores. This dynamically adaptive allocation strategy significantly improves both model performance and parameter efficiency, effectively reinforcing the VFM teacher's capacity for domain adaptation.

\subsection{SAM-based Pseudo-Label Generation}

Inspired by SAM \cite{DBLP:conf/iccv/KirillovMRMRGXW23} and its promptable segmentation using sparse points and low-resolution dense masks, we introduce SAM-PLG for sparse-label image segmentation. This innovative strategy addresses the challenge of weak supervision.

In the SAM predictor framework, each inference input is defined as a pair $(P_i, M_i) = ((X, Y), (H_{lr}\times W_{lr}))$, with the corresponding output comprising three valid masks. Here, $P_i$ represents the pixel coordinates of the prompt query, and $M_i$ denotes a low-resolution dense embedding mask at a quarter of the original resolution. In the dense mask, a positive value indicates the presence of a foreground pixel for the instance prompted by pixel $P_i$, while a negative value signifies a background pixel. The absolute value of each pixel's numerical value reflects the confidence level associated with its classification as foreground or background.

We initially generate sparse labels based on the correspondence between points and pixels. For efficiency, these labels are downsampled to 1/4 resolution and employed as prompt points after resizing to the shape $1024\times 1024$ specified by SAM. Furthermore, in pursuit of more precise masks, we generate sparse masks based on semantic and instance labels, providing supplementary guiding information. Moving on, we feed the image and output to SAM and remove low-quality masks based on the computed stability scores. Following SAM, mask boxes are computed, and box-based non-maximal suppression (NMS) is implemented to mitigate excessive mask overlaps. For each valid mask, we assign semantic and instance labels by associating it with the predominant ground truth category contained within. Finally, following the ascending order of predicted IoU scores, masks are progressively overlapped to yield the final high-quality pseudo-labels.

\section{Experiments}

\begin{table*}[htbp]
\caption{Ablation study on the SemanticKITTI validation split}
\setlength\tabcolsep{8pt}
\begin{center}
\begin{tabular}{c|c|c|c|c|c|c|c}
\toprule
\multirow{2}{*}{\textbf{No.}}
& \multirow{2}{*}{\textbf{Baseline}}
& \multicolumn{2}{c|}{\textbf{VFM Teacher}}
& \multicolumn{2}{c|}{\textbf{PPMSKD}}
& \multirow{2}{*}{\textbf{SAM-PLG}}
& \multirow{2}{*}{\textbf{mIoU(\%)}} \\

&
& Tuning Mode & Memory Optimization
& Patch Decoder Design & KD
&
& \\
\midrule
\emph{a} & \checkmark & & & & & & 62.6 \\
\emph{b} & \checkmark & Fixed & & MaxPool \& repeat & KD & & $63.5(0.9\uparrow)$ \\
\emph{c} & \checkmark & Fixed & & MaxPool \& repeat & MSKD & & $65.5(2.9\uparrow)$ \\
\emph{d} & \checkmark & Fixed & \checkmark & MaxPool \& repeat & MSKD & & $65.1(2.5\uparrow)$ \\
\midrule
\emph{e} & \checkmark & Fixed & & AvgPool \& repeat & MSKD & & $65.5(2.9\uparrow)$ \\
\emph{f} & \checkmark & Fixed & & neck \& repeat & MSKD & & $65.5(2.9\uparrow)$ \\
\emph{g} & \checkmark & Fixed & \checkmark & upscaling \& interpolate & MSKD & & $65.7(3.1\uparrow)$ \\
\midrule
\emph{h} & \checkmark & \makecell[c]{PE+LN Tuning*} & \checkmark & MaxPool \& repeat & MSKD & & $65.2(2.6\uparrow)$ \\
\emph{i} & \checkmark & LoRA & \checkmark & MaxPool \& repeat & MSKD & & $65.4(2.8\uparrow)$ \\
\emph{j} & \checkmark & AdaLoRA & \checkmark & MaxPool \& repeat & MSKD & & $65.5(2.9\uparrow)$ \\
\midrule
\emph{k} & \checkmark & AdaLoRA & \checkmark & upscaling \& interpolate & MSKD & & $66.2(3.6\uparrow)$ \\
\emph{l} & \checkmark & AdaLoRA & \checkmark & upscaling \& interpolate & MSKD & \checkmark & \textbf{66.4(3.8}$\uparrow$\textbf{)} \\
\bottomrule
\multicolumn{8}{l}{$^{\mathrm{*}}$PE+LN Tuning signifies only fine-tuning on Patch Embedding and Layer Normalization.}
\end{tabular}
\label{tab:ablation}
\end{center}
\end{table*}

\subsection{Dataset, Metrics, and Implementation Details}
We evaluate our approach using the large-scale outdoor benchmark SemanticKITTI \cite{DBLP:conf/iccv/BehleyGMQBSG19}. This dataset furnishes comprehensive semantic annotations for scans in sequences 00–10 of the KITTI dataset \cite{DBLP:conf/cvpr/GeigerLU12}. In line with official guidelines, sequence 08 constitutes the validation split, with the others forming the training split. For the test set, SemanticKITTI employs sequences 11–21 from the KITTI dataset, keeping labels concealed for blind online testing.

Our primary evaluation metric is the mean Intersection over Union (mIoU), which calculates the average IoU across all classes. We also assess the method's practicality by measuring its inference speed in frames per second (FPS). Training parameters are used as an indirect metric of training efficiency.
We employ the cross-entropy and Lovasz losses as \cite{DBLP:conf/eccv/XuGZZZCL22} for both image and LiDAR semantic segmentation.

We use a SAM pre-trained ViT base \cite{DBLP:conf/iclr/DosovitskiyB0WZ21} encoder as the VFM Teacher network, and the features output from each global attention block are selected for knowledge transfer.
% The LiDAR student encoder is a modified SPVCNN \cite{DBLP:conf/eccv/TangLZLLWH20} with few parameters, whose hidden dimensions are 64 to speed up the network and the number of layers is 4.

\subsection{Comparison Results}

Table \ref{tab:comparison} presents the performance under the single scan configuration. Compared to our baseline 2DPASS-base, ELiTe brings significant improvements (71.4 vs. 67.7). ELiTe enables the use of a student model with only 1.9M parameters for real-time inference (24Hz) and highly competitive performance (71.4). In comparison with real-time inference methods, our performance (71.4 vs. 71.2), speed (24Hz vs. 20Hz), and training efficiency are state-of-the-art, demonstrating the practicality and balance of ELiTe. Notably, our speed significantly surpasses other real-time inference methods (24Hz vs. 12-20Hz). Additionally, compared to non-real-time methods with a large number of parameters, we outperform them in speed (24Hz vs. 7Hz) and training efficiency (6.8M vs. 74.3M), remaining highly competitive in performance.

\subsection{Ablation Studies}

Table \ref{tab:ablation} summarizes ablation studies on the SemanticKITTI validation set. The student network baseline resulted in a relatively lower mIoU of 62.6 (\emph{a}). The use of vanilla knowledge distillation alone yielded marginal improvement, with a mere 0.9 increase in mIoU (\emph{b} vs. \emph{a}). PPMSKD introduced more efficient knowledge transfer, contributing an additional 2.0 improvement (\emph{c} vs. \emph{a}). Additionally, for memory conservation, certain patch-level padding features were discarded, causing a minor performance dip of 0.4 (\emph{d} vs. \emph{c}).

In the subsequent comparisons, we explore effective Patch Encoder forms to optimize PPMSKD. Finally, by employing a feature dimension reduction and downsampling approach similar to SAM's decoder, using an upscale-interpolate method, we achieved superior results (\emph{g} vs. \emph{c, e, f}). This outcome was attained while adhering to memory optimization considerations (\emph{g} vs. \emph{d}).

Due to memory and computational constraints, we employed three fine-tuning methods rather than fully training the teacher model. These methods improved model performance (\emph{h, i, j} vs. \emph{d}), including partial teacher layer fine-tuning, LoRA, and AdaLoRA. Importantly, it is observed that LoRA-based approaches yielded the most favorable results (\emph{i, j} vs. \emph{h}).

In the end, with optimal PPMSKD and VFM training, we achieved a notable score of 66.2 (\emph{k} vs. \emph{g, j}). Additionally, the integration of pseudo-labels SAM-PLG generated further enhanced performance (\emph{l} vs. \emph{k}), affirming the positive impact of strong teacher supervision on the student's progress.

\subsection{Comprehensive Analysis}

\begin{table}[htbp]
\caption{Comprehensive study on teacher modality choice}
\begin{center}
\begin{tabular}{c|c|c|c}
\toprule
\textbf{Method}
& \textbf{Teacher Modality}
& \textbf{Teacher mIoU}
& \textbf{Student mIoU} \\
\midrule
2DPASS & 2D \& 3D & 64.6 & 65.4\\
2DPASS & Pure 2D & 21.5 & 65.2\\
ELiTe & Pure 2D & 34.0 & \textbf{66.2} \\
ELiTe & 2D \& 3D & \textbf{64.9} & 65.5 \\
\bottomrule
\end{tabular}
\label{tab:teacher}
\end{center}
\end{table}

In evaluating our approach to the teacher network, assessments were conducted using both pure 2D and fused-modal 2DPASS \cite{DBLP:conf/eccv/XuGZZZCL22}, along with the ELiTe methods. Results in Table \ref{tab:teacher} show that while fused teachers outperformed pure 2D counterparts (64.6 vs. 21.5), their actual benefit to the student network was limited (65.4 vs. 65.2). In the context of 2DPASS, due to the complexity of extracting effective 2D features, fused teacher features gradually resembled those of the LiDAR student network after attention-based weighting. This inadvertently led to a self-distillation loop, posing a challenge for the student to imitate image-specific features.

Conversely, within the ELiTe framework, although the performance of the optimal VFM teacher still notably lags behind that of the student (34.0 vs. 64.6), the student's performance shows improvement compared to 2DPASS(66.2 vs. 65.4). This effectively underscores a concept: \textbf{the student needs not only a brilliant teacher but also one who has unique and new knowledge.} Furthermore, by incorporating fused teachers into ELiTe, its results are similar to vanilla 2DPASS (64.9 \& 65.5 vs. 64.6 \& 65.4), this principle is reiterated, underscoring that the top-performing teacher does not necessarily guarantee the best student performance.

\section{Conclusion}

In this study, we present ELiTe, an Efficient Image-to-LiDAR Knowledge Transfer paradigm. ELiTe effectively addresses weak teacher model challenges in cross-modal knowledge transfer for LiDAR semantic segmentation with significantly few parameters. It combines VFM Teacher, PPMSKD, PEFT, and SAM-PLG methods to optimize teacher model encoding, promote domain adaptation, facilitate knowledge transfer to a lightweight student model, and employ a pioneering pseudo-label generation strategy to enhance semantic representations' robustness. Through comprehensive SemanticKITTI benchmark evaluation, ELiTe demonstrates superior real-time inference performance, showcasing the potential for enhancing LiDAR-based perception across applications.

\appendix

In this appendix, we present additional details of implementations, experiments, and visualizations for a better understanding.

\subsection{SAM-PLG}
We demonstrate the specific automatic generation process in Algorithm Stage \ref{alg:alg1} and \ref{alg:alg2}. In Stage \ref{alg:alg1}, we initially generate sparse labels based on the correspondence between points and pixels. For efficiency, these labels are downsampled to 1/4 resolution and employed as prompt points after resizing to the shape specified by SAM ($1024\times 1024$). Furthermore, in pursuit of more precise masks, we generate sparse masks based on semantic and instance labels, providing supplementary guiding information. 

Moving on to Stage \ref{alg:alg2}, we feed the image and output of Stage \ref{alg:alg1} to SAM and remove low-quality masks based on the computed stability scores. Following SAM, mask boxes are computed, and box-based non-maximal suppression (NMS) is implemented to mitigate excessive mask overlaps. For each valid mask, we assign semantic and instance labels by associating it with the predominant ground truth category contained within. Finally, following the ascending order of predicted IoU scores, masks are progressively overlapped to yield the final high-quality pseudo-labels.

\begin{algorithm}[htbp]
\caption{SAM-based Pseudo-Label Generation (Stg. 1)}
\label{alg:alg1}
\textbf{Input}: Sparse semantic level label $L_{s}\in R^{H\times W}$, instance level label $L_{i}\in R^{H\times W}$\\
\textbf{Parameter}: The low resolution size $[H_{lr}, W_{lr}]$, high and low confidence score $\theta_h,\theta_l$ for dense mask\\
\textbf{Output}: Point prompts $P\in R^{N_{lr}\times 2}$, low resolution mask input $M\in R^{N_{lr}\times H_{lr}\times W_{lr}}$
\begin{algorithmic}[1] %[1] enables line numbers
\STATE Let $L_{s}^{lr}, L_{i}^{lr}\leftarrow downsampling(L_{s}, L_{i})$;
\STATE Let $P\leftarrow where(L_{s}^{lr}$ is not ignore label$)$;
\STATE Let $N_{lr}\leftarrow len(P)$;
\STATE Let $M\leftarrow zeros([N_{lr}, H_{lr}, W_{lr}])$;
\FOR{$i$ in $range(N_{lr})$}
\IF{$L_{s}^{lr}[P[i]]$ is not ignore label}
\STATE $M[i][L_{s}^{lr}[P] \neq L_{s}^{lr}[P[i]]] \leftarrow -\theta_h$;
\IF{$L_{i}^{lr}[P[i]]$ is valid}
\STATE $M[i][L_{i}^{lr}[P] = L_{i}^{lr}[P[i]]] \leftarrow \theta_h$;
\ELSE
\STATE $M[i][L_{s}^{lr}[P] = L_{s}^{lr}[P[i]]] \leftarrow \theta_l$;
\ENDIF
\ENDIF
\ENDFOR
\STATE \textbf{return} $P$, $M$
\end{algorithmic}
\end{algorithm}

\begin{algorithm}[htbp]
\caption{SAM-based Pseudo-Label Generation (Stg. 2)}
\label{alg:alg2}
\textbf{Input}: Car camera image $I\in R^{H\times W\times rgb}$, point prompts $P$, low resolution mask input $M$\\
\textbf{Parameter}: The SAM model $SAM$, the stability score filtering threshold $\theta_{stability}$, the box IoU cutoff used by non-maximal suppression $\theta_{box\_nms}$\\
\textbf{Output}: Semantic pseudo-label $PL_{s}\in R^{H\times W}$, instance pseudo-label $PL_{i}\in R^{H\times W}$
\begin{algorithmic}[1] %[1] enables line numbers
\STATE Let $data\leftarrow \{\}$;
\FOR{$i$ in $range(N_{lr})$}
\STATE $masks, iou\_predictions\leftarrow SAM(I,P[i],M[i])$;
\STATE $S_{stability}\leftarrow calculate\_stability\_score(masks)$;
\FOR{$mask$ in $masks$}
\IF{$S_{stability}> \theta_{stability}$}
\STATE $box \leftarrow mask\_to\_box(mask)$
\STATE $data += \{mask, iou\_prediction, box\}$
\ENDIF
\ENDFOR
\ENDFOR
\STATE $data \leftarrow non\_maximum\_suppression(data,\theta_{box\_nms})$
\STATE Sort $data$ in ascending order by $iou\_prediction$
\STATE Let $PL_{s},PL_{i}\leftarrow zeros([H, W]),zeros([H, W])$;
\FOR{$d$ in $data$}
\STATE $PL_{s}[d[mask]]\leftarrow most\_common(L_{s}[d[mask]])$;
\STATE $PL_{i}[d[mask]]\leftarrow most\_common(L_{i}[d[mask]])$;
\ENDFOR
\STATE \textbf{return} $PL_{s}, PL_{i}$
\end{algorithmic}
\end{algorithm}

\subsection{Implementation Details}

\subsubsection{Lightweight Student}
The LiDAR student encoder is a modified SPVCNN \cite{DBLP:conf/eccv/TangLZLLWH20} with few parameters, whose hidden dimensions are 64 to speed up the network and the number of layers is 4.

\subsubsection{Loss Function}
We employ the cross-entropy and Lovasz losses as \cite{DBLP:conf/eccv/XuGZZZCL22} for both 2D and 3D semantic segmentation: $L_{seg}=L_{wce}+L_{lovasz}$. Due to the utilization of segmentation heads in both the teacher and student networks, a total of $2L+2$ segmentation losses are employed. Among them, the weights of the $2L$ single-stage losses are uniformly set to $1/L$. Apart from the segmentation and distillation losses, an additional loss for computing orthogonal regularization is incorporated to enhance the AdaLoRA optimization process.

\subsubsection{Training and Inference}
Only 64 training epochs were used in the ablation experiment and comprehensive analysis. Test-time augmentation is only applied during the inference on test split.

\subsubsection{SAM-PLG}
In SAM-PLG, we adopted the pre-trained parameters from SAM with ViT-Huge. The low-resolution size for dense masks is set at [256, 256], with high and low confidence scores $\theta_h,\theta_l$ of 16 and 1, respectively. The stability score filtering threshold $\theta_{stability}$ is set at 0.9, and the non-maximal suppression threshold $\theta_{box\_nms}$ is set at 0.7.

\subsection{More Experiment Details}

\subsubsection{Comparison Results}
The segmentation performance metrics in the "Comparison Results" section are sourced from the official SemanticKITTI \cite{DBLP:conf/iccv/BehleyGMQBSG19} website. However, the exception is the "2DPASS-base" results, which are obtained using the provided open-source checkpoint following official settings.

For the FPS (frames per second) speed metric, testing was conducted on a single NVIDIA GeForce RTX 3090 card on the validation set without using batch size. The $torch.cuda.synchronize()$ function was used during testing to ensure accuracy.

\subsubsection{Parameters Details}

\begin{table*}[htbp]
\caption{Detailed parameters distribution}
\centering
\begin{tabular}{c|c|c|c|c}
\toprule
\textbf{Params Type}
& \textbf{Lightweight Student}
& \textbf{VFM Teacher}
& \textbf{PPMSKD}
& \textbf{Total}\\
\midrule
Trainable & 1.9M & 2.0M & 2.9M & 6.8M\\
Non-trainable & 0 & 88.9M & 0 & 88.9M\\
Total & 1.9M & 90.9M & 2.9M & 95.7M\\
\bottomrule
\end{tabular}
\label{tab:params}
\end{table*}

For the model parameters, some values are taken from the results reported, while others are derived from the official open-source code's models. The specific calculation method for the total parameters is:
$$sum([x.nelement()\ for\ x\ in\ model.parameters()]).$$

As shown in Table \ref{tab:params}, we present a more detailed depiction of the parameter distribution for ELiTe. Besides, in Table 1 of the main content, it is noted that the training parameters for SPVNAS and Point-Voxel-KD might exceed the inference parameters, hence they have not been recorded.

\subsection{More Visualization Details}

\subsubsection{Car Camera Images and Open World Images}

The "Open World Images" is sourced from the SAM \cite{DBLP:conf/iccv/KirillovMRMRGXW23} demo. The "Car Camera Images" are derived from the KITTI \cite{DBLP:conf/cvpr/GeigerLU12} dataset's sequence 08, comprising frames numbered 000000-000002, 001000-001002, 002000-002002, and 003000-003002. For ease of viewing, these images have undergone cropping and stretching. The sequential car images exhibit significant redundancy across adjacent frames, resulting in a lack of overall diversity and a monotonous style.

\subsubsection{Feature Visualization}

\begin{figure*}[htbp]
\centering
\includegraphics[width=\textwidth]{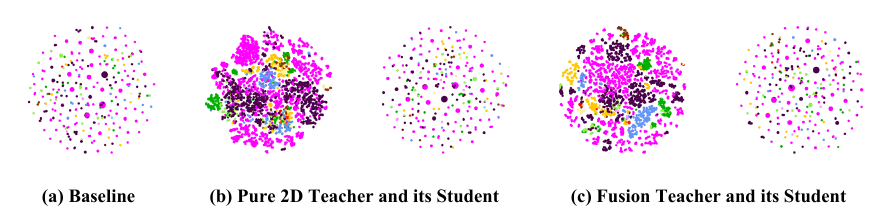}
\caption{T-SNE Visualization. In (b) and (c), the left part clusters are from the image teacher, and the right part clusters are from the LiDAR student. The feature is the intersection of points and pixels extracted from a single frame segmentation, so each group of clusters has the same number of feature points.}
\label{fig_fea}
\end{figure*}

We utilize t-SNE to visualize features from the image teacher and LiDAR student. T-SNE effectively captures neural network clustering patterns, as depicted in Figure \ref{fig_fea}(b). Notably, LiDAR student features form distinct compact clusters, while image segmentation teacher's features align more closely with conventional image classification features \cite{DBLP:conf/eccv/XuYGWYZZVKT22}, resulting in expansive and continuous clusters.

We speculate that the sparse clustering of LiDAR student features could be attributed to constrained data diversity and weaker model generalization, thus avoiding the conventional neural collapse phenomena \cite{DBLP:conf/iclr/Galanti0H22}. The distinct clustering pattern of features emphasizes how the inclusion of image teachers provides distinct image features for the students to emulate. However, it's important to note that the student's feature clustering does not precisely replicate the teacher's, suggesting that traditional knowledge distillation might not facilitate effective cross-modal transfer. This underscores a limitation in our approach.

In Figure \ref{fig_fea}(a)(c), we further employ t-SNE to visualize features from the baseline, 2D, and 3D fused teacher, along with its corresponding student. It is noteworthy that the features from the LiDAR baseline or student consistently form distinct and compact clusters, while the teacher's features exhibit a closer resemblance to conventional image classification features \cite{DBLP:conf/eccv/XuYGWYZZVKT22}, displaying relatively continuous clustering patterns. These observations yield the same conclusion above: the student's features do not accurately mimic the teacher's clustering, indicating that vanilla knowledge distillation still exhibits shortcomings in cross-modal knowledge transfer.

\subsubsection{Pseudo-Label Visualization.}

\begin{figure}[htbp]
\centering
\includegraphics[width=\columnwidth]{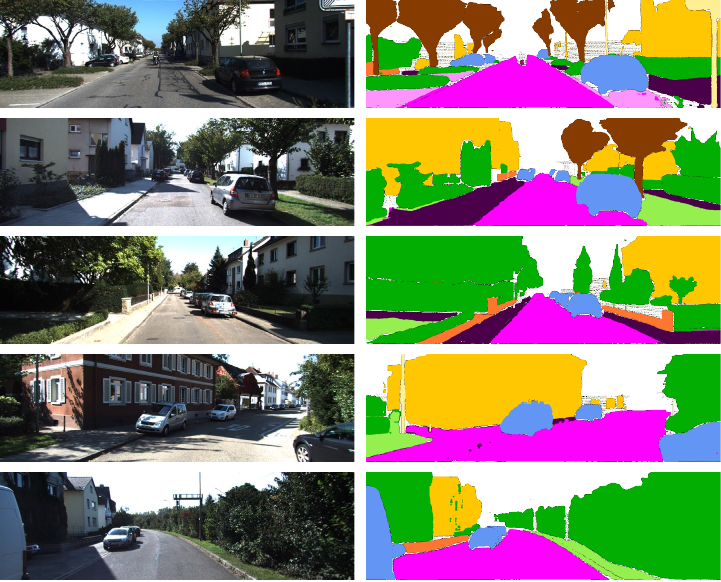}
\caption{Visualization of images and their Pseudo-Labels.}
\label{fig5}
\end{figure}

We illustrate SAM-PLG's excellence through Figure \ref{fig5}, highlighting our achievement in generating comprehensive and dense pseudo labels. However, we acknowledge sporadic gaps in pseudo-label density and occasional imprecision in segmentation boundaries. These limitations stem from both the inherent inaccuracies within the initial ground truth labels we rely on, and our strategic trade-off between efficiency and absolute precision.

The images are sourced from the KITTI \cite{DBLP:conf/cvpr/GeigerLU12} dataset sequence 00, encompassing frames with sequential numbers 000000, 001000, 002000, 003000, and 004000. Randomly selected images exhibit high-quality pseudo-labels, effectively encompassing the majority of primary categories and excluding out-of-domain classes such as the sky.

\subsubsection{Result Visualization}

\begin{figure*}[htbp]
\centering
\includegraphics[width=\textwidth]{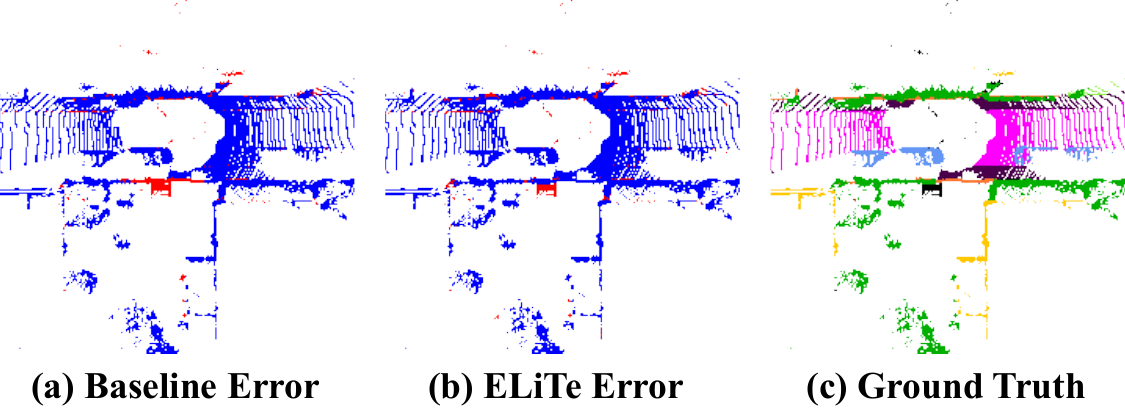}
\caption{More Result Visualization.}
\label{fig_result}
\end{figure*}

In Figure \ref{fig_result},  we provide a bird's-eye view of the resulting error between baseline and ELiTe comparison. ELiTe shows less error at close range.

\subsection{More Related Work}

\subsubsection{Foundation Models}
Pre-trained models have been adapted to downstream tasks since the early days of machine learning. This paradigm has become increasingly important in recent years with a growing emphasis on scale, and such models trained on broad data at scale have recently been re-branded as foundation models. The widely known foundation model is the \textbf{Large Language Model (LLM)} in Natural Language Processing (NLP), such as GPT-4 \cite{achiam2023gpt} and LLaMA \cite{touvron2023llama}. Recently, the concept of \textbf{Vision Foundation Model (VFM)} has also emerged in computer vision. Among them, the Segment Anything Model (SAM) \cite{DBLP:conf/iccv/KirillovMRMRGXW23} is renowned for its powerful image zero-shot instance segmentation. With the emergence of SAM, there have been many attempts \cite{zhang2023sam3d} \cite{yang2023sam3d} \cite{DBLP:conf/nips/CenZF00XJZ023} to use it for 3D understanding. They have all achieved remarkable results in their respective fields, but they mainly focus on the results of using SAM directly, rather than incorporating SAM into the training process.

\bibliographystyle{IEEEbib}
\bibliography{IEEEabrv}

\begin{thebibliography}{10}

\bibitem{DBLP:journals/tits/GaoPLGZ22}
Biao Gao, Yancheng Pan, Chengkun Li, Sibo Geng, and Huijing Zhao,
\newblock ``Are we hungry for 3d lidar data for semantic segmentation? {A} survey of datasets and methods,''
\newblock {\em {IEEE} Trans. Intell. Transp. Syst.}, vol. 23, no. 7, pp. 6063--6081, 2022.

\bibitem{DBLP:conf/eccv/XuYGWYZZVKT22}
Chenfeng Xu, Shijia Yang, Tomer Galanti, Bichen Wu, Xiangyu Yue, Bohan Zhai, Wei Zhan, Peter Vajda, Kurt Keutzer, and Masayoshi Tomizuka,
\newblock ``Image2point: 3d point-cloud understanding with 2d image pretrained models,''
\newblock in {\em ECCV}, 2022, vol. 13697, pp. 638--656.

\bibitem{DBLP:conf/eccv/XuGZZZCL22}
Yan Xu, Jiantao Gao, Chaoda Zheng, Chao Zheng, Ruimao Zhang, Shuguang Cui, and Zhen Li,
\newblock ``2dpass: 2d priors assisted semantic segmentation on lidar point clouds,''
\newblock in {\em ECCV}, 2022, vol. 13688, pp. 677--695.

\bibitem{DBLP:conf/aaai/XingYWYC23}
Bowei Xing, Xianghua Ying, Ruibin Wang, Jinfa Yang, and Taiyan Chen,
\newblock ``Cross-modal contrastive learning for domain adaptation in 3d semantic segmentation,''
\newblock in {\em AAAI}, 2023, pp. 2974--2982.

\bibitem{DBLP:conf/iccv/BehleyGMQBSG19}
Jens Behley, Martin Garbade, Andres Milioto, Jan Quenzel, Sven Behnke, Cyrill Stachniss, and J{\"{u}}rgen Gall,
\newblock ``Semantickitti: {A} dataset for semantic scene understanding of lidar sequences,''
\newblock in {\em ICCV}, 2019, pp. 9296--9306.

\bibitem{DBLP:conf/iccv/KirillovMRMRGXW23}
Alexander Kirillov, Eric Mintun, Nikhila Ravi, Hanzi Mao, Chlo{\'{e}} Rolland, Laura Gustafson, Tete Xiao, Spencer Whitehead, Alexander~C. Berg, Wan{-}Yen Lo, Piotr Doll{\'{a}}r, and Ross~B. Girshick,
\newblock ``Segment anything,''
\newblock in {\em ICCV}, 2023, pp. 3992--4003.

\bibitem{DBLP:conf/cvpr/GeigerLU12}
Andreas Geiger, Philip Lenz, and Raquel Urtasun,
\newblock ``Are we ready for autonomous driving? the {KITTI} vision benchmark suite,''
\newblock in {\em CVPR}, 2012, pp. 3354--3361.

\bibitem{DBLP:conf/cvpr/HouZMLL22}
Yuenan Hou, Xinge Zhu, Yuexin Ma, Chen~Change Loy, and Yikang Li,
\newblock ``Point-to-voxel knowledge distillation for lidar semantic segmentation,''
\newblock in {\em CVPR}, 2022, pp. 8469--8478.

\bibitem{DBLP:conf/cvpr/Zhu0WHM00L21}
Xinge Zhu, Hui Zhou, Tai Wang, Fangzhou Hong, Yuexin Ma, Wei Li, Hongsheng Li, and Dahua Lin,
\newblock ``Cylindrical and asymmetrical 3d convolution networks for lidar segmentation,''
\newblock in {\em CVPR}, 2021, pp. 9939--9948.

\bibitem{DBLP:conf/iccv/XuZDZSP21}
Jianyun Xu, Ruixiang Zhang, Jian Dou, Yushi Zhu, Jie Sun, and Shiliang Pu,
\newblock ``Rpvnet: {A} deep and efficient range-point-voxel fusion network for lidar point cloud segmentation,''
\newblock in {\em ICCV}, 2021, pp. 16004--16013.

\bibitem{DBLP:conf/iclr/HuSWALWWC22}
Edward~J. Hu, Yelong Shen, Phillip Wallis, Zeyuan Allen{-}Zhu, Yuanzhi Li, Shean Wang, Lu~Wang, and Weizhu Chen,
\newblock ``Lora: Low-rank adaptation of large language models,''
\newblock in {\em ICLR}, 2022.

\bibitem{DBLP:conf/eccv/TangLZLLWH20}
Haotian Tang, Zhijian Liu, Shengyu Zhao, Yujun Lin, Ji~Lin, Hanrui Wang, and Song Han,
\newblock ``Searching efficient 3d architectures with sparse point-voxel convolution,''
\newblock in {\em ECCV}, 2020, vol. 12373, pp. 685--702.

\bibitem{DBLP:conf/itsc/MadawiRSNKY19}
Khaled~El Madawi, Hazem Rashed, Ahmad~El Sallab, Omar Nasr, Hanan Kamel, and Senthil~Kumar Yogamani,
\newblock ``{RGB} and lidar fusion based 3d semantic segmentation for autonomous driving,''
\newblock in {\em ITSC}, 2019, pp. 7--12.

\bibitem{DBLP:conf/iccv/ZhuangLJWLT21}
Zhuangwei Zhuang, Rong Li, Kui Jia, Qicheng Wang, Yuanqing Li, and Mingkui Tan,
\newblock ``Perception-aware multi-sensor fusion for 3d lidar semantic segmentation,''
\newblock in {\em ICCV}, 2021, pp. 16260--16270.

\bibitem{DBLP:conf/cvpr/DengDSLL009}
Jia Deng, Wei Dong, Richard Socher, Li{-}Jia Li, Kai Li, and Li~Fei{-}Fei,
\newblock ``Imagenet: {A} large-scale hierarchical image database,''
\newblock in {\em CVPR}, 2009, pp. 248--255.

\bibitem{DBLP:conf/cvpr/ZhangGZLM0QG022}
Renrui Zhang, Ziyu Guo, Wei Zhang, Kunchang Li, Xupeng Miao, Bin Cui, Yu~Qiao, Peng Gao, and Hongsheng Li,
\newblock ``Pointclip: Point cloud understanding by {CLIP},''
\newblock in {\em CVPR}, 2022, pp. 8542--8552.

\bibitem{DBLP:conf/iclr/DosovitskiyB0WZ21}
Alexey Dosovitskiy, Lucas Beyer, Alexander Kolesnikov, Dirk Weissenborn, Xiaohua Zhai, Thomas Unterthiner, Mostafa Dehghani, Matthias Minderer, Georg Heigold, Sylvain Gelly, Jakob Uszkoreit, and Neil Houlsby,
\newblock ``An image is worth 16x16 words: Transformers for image recognition at scale,''
\newblock in {\em ICLR}, 2021.

\bibitem{DBLP:conf/nips/VaswaniSPUJGKP17}
Ashish Vaswani, Noam Shazeer, Niki Parmar, Jakob Uszkoreit, Llion Jones, Aidan~N. Gomez, Lukasz Kaiser, and Illia Polosukhin,
\newblock ``Attention is all you need,''
\newblock in {\em Advances in Neural Information Processing Systems}, 2017, pp. 5998--6008.

\bibitem{DBLP:conf/aaai/ZhangYWJ22}
Kaiyi Zhang, Ximing Yang, Yuan Wu, and Cheng Jin,
\newblock ``Attention-based transformation from latent features to point clouds,''
\newblock in {\em AAAI}, 2022, pp. 3291--3299.

\bibitem{DBLP:conf/mm/CaoZCYJ22}
Rui Cao, Kaiyi Zhang, Yang Chen, Ximing Yang, and Cheng Jin,
\newblock ``Point cloud completion via multi-scale edge convolution and attention,''
\newblock in {\em MM}, 2022, pp. 6183--6192.

\bibitem{DBLP:conf/icmcs/WangFWXZ23}
Yibin Wang, Yuchao Feng, Jie Wu, Honghui Xu, and Jianwei Zheng,
\newblock ``{CA-GAN:} object placement via coalescing attention based generative adversarial network,''
\newblock in {\em ICME}, 2023, pp. 2375--2380.

\bibitem{wang2024residual}
Yibin Wang, Haixia Long, Tao Bo, and Jianwei Zheng,
\newblock ``Residual graph transformer for autism spectrum disorder prediction,''
\newblock {\em Computer Methods and Programs in Biomedicine}, p. 108065, 2024.

\bibitem{yang2021generate}
Ximing Yang, Zhibo Zhang, Zhengfu He, and Cheng Jin,
\newblock ``Generate point clouds with multiscale details from graph-represented structures,''
\newblock {\em arXiv preprint arXiv:2112.06433}, 2021.

\bibitem{DBLP:conf/icra/GerdzhevRTL21}
Martin Gerdzhev, Ryan Razani, Ehsan Taghavi, and Bingbing Liu,
\newblock ``Tornado-net: multiview total variation semantic segmentation with diamond inception module,''
\newblock in {\em ICRA}, 2021, pp. 9543--9549.

\bibitem{DBLP:conf/cvpr/ChengRTLL21}
Ran Cheng, Ryan Razani, Ehsan Taghavi, Enxu Li, and Bingbing Liu,
\newblock ``(af)2-s3net: Attentive feature fusion with adaptive feature selection for sparse semantic segmentation network,''
\newblock in {\em CVPR}, 2021, pp. 12547--12556.

\bibitem{DBLP:conf/iros/MiliotoVBS19}
Andres Milioto, Ignacio Vizzo, Jens Behley, and Cyrill Stachniss,
\newblock ``Rangenet ++: Fast and accurate lidar semantic segmentation,''
\newblock in {\em IROS}, 2019, pp. 4213--4220.

\bibitem{DBLP:conf/cvpr/0035ZDYXGF20}
Yang Zhang, Zixiang Zhou, Philip David, Xiangyu Yue, Zerong Xi, Boqing Gong, and Hassan Foroosh,
\newblock ``Polarnet: An improved grid representation for online lidar point clouds semantic segmentation,''
\newblock in {\em CVPR}, 2020, pp. 9598--9607.

\bibitem{DBLP:conf/icra/RazaniCTL21}
Ryan Razani, Ran Cheng, Ehsan Taghavi, and Bingbing Liu,
\newblock ``Lite-hdseg: Lidar semantic segmentation using lite harmonic dense convolutions,''
\newblock in {\em ICRA}, 2021, pp. 9550--9556.

\bibitem{DBLP:conf/cvpr/WortsmanIKLKRLH22}
Mitchell Wortsman, Gabriel Ilharco, Jong~Wook Kim, Mike Li, Simon Kornblith, Rebecca Roelofs, Raphael~Gontijo Lopes, Hannaneh Hajishirzi, Ali Farhadi, Hongseok Namkoong, and Ludwig Schmidt,
\newblock ``Robust fine-tuning of zero-shot models,''
\newblock in {\em CVPR}, 2022, pp. 7949--7961.

\bibitem{zhang2024robust}
Zhibo Zhang, Ximing Yang, Weizhong Zhang, and Cheng Jin,
\newblock ``Robust fine-tuning for pre-trained 3d point cloud models,''
\newblock {\em arXiv preprint arXiv:2404.16422}, 2024.

\bibitem{zhang2023adaptive}
Qingru Zhang, Minshuo Chen, Alexander Bukharin, Pengcheng He, Yu~Cheng, Weizhu Chen, and Tuo Zhao,
\newblock ``Adaptive budget allocation for parameter-efficient fine-tuning,''
\newblock in {\em ICLR}, 2023.

\bibitem{DBLP:conf/iclr/Galanti0H22}
Tomer Galanti, Andr{\'{a}}s Gy{\"{o}}rgy, and Marcus Hutter,
\newblock ``On the role of neural collapse in transfer learning,''
\newblock in {\em ICLR}, 2022.

\bibitem{achiam2023gpt}
Josh Achiam, Steven Adler, Sandhini Agarwal, Lama Ahmad, Ilge Akkaya, Florencia~Leoni Aleman, Diogo Almeida, Janko Altenschmidt, Sam Altman, Shyamal Anadkat, et~al.,
\newblock ``Gpt-4 technical report,''
\newblock {\em arXiv preprint arXiv:2303.08774}, 2023.

\bibitem{touvron2023llama}
Hugo Touvron, Thibaut Lavril, Gautier Izacard, Xavier Martinet, Marie-Anne Lachaux, Timoth{\'e}e Lacroix, Baptiste Rozi{\`e}re, Naman Goyal, Eric Hambro, Faisal Azhar, et~al.,
\newblock ``Llama: Open and efficient foundation language models,''
\newblock {\em arXiv preprint arXiv:2302.13971}, 2023.

\bibitem{zhang2023sam3d}
Dingyuan Zhang, Dingkang Liang, Hongcheng Yang, Zhikang Zou, Xiaoqing Ye, Zhe Liu, and Xiang Bai,
\newblock ``Sam3d: Zero-shot 3d object detection via segment anything model,''
\newblock {\em arXiv preprint arXiv:2306.02245}, 2023.

\bibitem{yang2023sam3d}
Yunhan Yang, Xiaoyang Wu, Tong He, Hengshuang Zhao, and Xihui Liu,
\newblock ``Sam3d: Segment anything in 3d scenes,''
\newblock {\em arXiv preprint arXiv:2306.03908}, 2023.

\bibitem{DBLP:conf/nips/CenZF00XJZ023}
Jiazhong Cen, Zanwei Zhou, Jiemin Fang, Chen Yang, Wei Shen, Lingxi Xie, Dongsheng Jiang, Xiaopeng Zhang, and Qi~Tian,
\newblock ``Segment anything in 3d with nerfs,''
\newblock in {\em NeurIPS}, 2023.

\end{thebibliography}

\end{document}